# TEXT MINING CUSTOMER REVIEWS FOR ASPECT-BASED RESTAURANT RATING

Jovelyn C. Cuizon , Jesserine Lopez and Danica Rose Jones

University of San Jose-Recoletos, Cebu City, Cebu Philippines

## ABSTRACT

*This study applies text mining to analyze customer reviews and automatically assign a collective restaurant star rating based on five predetermined aspects: ambiance, cost, food, hygiene, and service. The application provides a web and mobile crowd sourcing platform where users share dining experiences and get insights about the strengths and weaknesses of a restaurant through user contributed feedback. Text reviews are tokenized into sentences. Noun-adjective pairs are extracted from each sentence using Stanford Core NLP library and are associated to aspects based on the bag of associated words fed into the system. The sentiment weight of the adjectives is determined through AFINN library. An overall restaurant star rating is computed based on the individual aspect rating. Further, a word cloud is generated to provide visual display of the most frequently occurring terms in the reviews. The more feedbacks are added the more reflective the sentiment score to the restaurants' performance.*

## KEYWORDS

*Text Mining, Sentiment Analysis, Natural Language Processing, Aspect-based scoring*

## 1. INTRODUCTION

Customer feedbacks are useful for firms in order for them to recognize its strengths and weaknesses, and therefore generate ideas to improve its services. The proliferation of a wide variety of communication media has provided customers the capability to write and express their experiences about the products and services availed. Crowdsourcing feedback gives the customers the power to influence prospective customer's decision to avail of the products and services offered. Crowdsourcing applications have gained a lot of attention because it harnesses the potential of diverse group of people to provide information through various media. Zomato and Yelp are some of the many available crowdsourcing applications that gather customer feedback on restaurants. However, customer reviews come in bulk of unstructured text data that people need to read to understand the general perception of the customer on a restaurant. Customers are usually asked to assign a star rating in the range of 1 to 5 to assess the overall experience which may not necessary reflect the opinion in the textual feedback. The application of text mining for analysis of customer textual reviews to quantify it through star rating based on predetermined decision factors prove to be beneficial to help cope with the information overload and facilitate decision making.

Text mining encompasses varied techniques to analyze and digest information from textual data including natural language processing, information retrieval, data mining and machine learning [1]. Customer reviews served as corpus to understand the general perception of the customer towards the products and services. A study by Jack and Tsai relied on term frequency (n-gram) to aggregate top attributes and issues associated with devices such as laptops and tablets expressed in customer reviews to understand what customers liked or disliked about the products. [2] Ordenes et.al applied linguistic-based approach to evaluate the value creation components in customer feedback which influences customer experience [3]. Suresh et. al applied aspect-based opinion mining to recommend related restaurant reviews filtered according to predetermined features [4]. Hu et. al [5] and Somprasertsri et. al [6] applied text summarization to mine product features and opinions.

While there has been significant amount of study on text mining and sentiment analysis to understand customer reviews, converting textual data to numeric assessment to reflect overall perception of customer has not been extensively explored. System-assigned star rating minimizes if not eliminates inconsistency in the opinion expressed in text and the user-assigned numeric assessment.

This study aims to develop a mobile and web application that serves as a platform for diners to write feedback on dining experience. The system uses these reviews as corpus to determine customer perception on the restaurant in general and on the specific aspects such as ambience, cost, food, hygiene and service. A word cloud of the customers' general sentiment will give the restaurant the visual illustration of top qualities and issues.

## 2. RELATED WORKS

In order to understand the underlying meaning of a given text, text analysis algorithms are applied which enabled users to rapidly transform the key content in text documents into quantitative, actionable insights. Text mining encompasses techniques in data mining, information retrieval and natural language processing.

There have been a generous amount of studies on text mining for analysis of customer feedback or reviews. The study conducted by [7] and [8] incorporates the use of natural language processing to extract noun and adjective pairs from sentences through Parts-of-speech (POS) tagging and association rule mining on customer reviews of products to find frequent and infrequent features to ascertain product characteristics. Another study by [9] performed sentiment analysis and linguistic rules to analyze reviews and detect opinion orientation and important aspects about a restaurant.

A model proposed by [10] called Multi-Aspect Sentiment (MAS) model to discover topics in customer reviews and extract fragments of text that correspond to rateable aspect to support numerical ratings. An unsupervised method proposed by [11] extracts important aspects of a product to estimate an aspect rating from 1 to 5 to represent overall customer satisfaction.

## 3. PROPOSED WORK

As shown in Figure 1, the application runs in mobile and web platforms. The mobile component is developed using Android. The web application is developed using Java Platform, Enterprise Edition (J2EE) with Maven as a build automation tool and utilizes Spring for MVC (Model-View-Controller) framework for implementation. An application server is implemented which manages process requests from and to the mobile and web client applications through RESTful web service calls. To ensure availability of data, Google app engine is used as cloud data store and for push notification services.

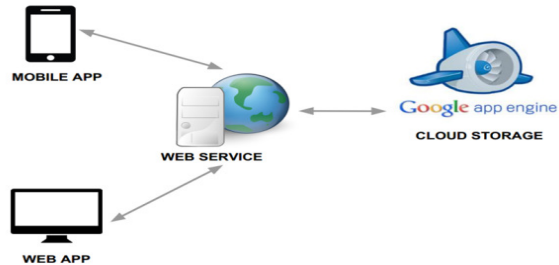

Figure 1  Architectural Diagram

A number of libraries and APIs are used to support in the development. Stanford Core NLP API is used to perform parts-of-speech (POS) tagging and noun-adjective pair extraction in sentences. AFINN library is utilized for determining polarity of words. Google Maps API, an Application Programming Interface is used to embed Google Maps.

### 3.1 Noun-Adjective Pairs Extraction

Automatic assignment of feature assessment rating is performed through the application of natural language processing (NLP) techniques on text reviews. Customer reviews can range from one phrase to sentences to paragraphs.  A sample of a review is as follows:

"This is one of the best places to eat lechon. The place is clean, the staff is friendly, and they have a menu that is filled with dishes that go so well with Cebu's Lechon. I usually take out of towners here when they crave for lechon and so far, all of them were happy with this place. Among my favorites here is their Carcar Lechon."

The text goes through sentence segmentation as shown in Figure 2 which generates a list of text-sentences, one in which that begins with capital letter and ends with a boundary punctuation marks. Sentence boundary punctuations included the period, question mark and exclamation point.

```
run:
Loading parser from serialized file edu/stanford/nlp/models/lexparser/englishPCFG.ser.gz
1       This is one of the best places to eat lechon
2       The place is clean, the staff is friendly and they have a menu that is filled wit
3       I usually take out of towners here when they crave for lechon and so far, all of
4       Among my favorites here is their Carcar Lechon
```

Figure 2 List of sentences after segmentation

Extracted sentences are passed to the Stanford CoreNLP API to retrieve a tree of dependencies in the sentence which is used to extract the noun-adjective (NA) pairs in the sentence.

```
run:
Loading parser from serialized file edu/stanford/nlp/models/lexparser/englishPCFG.ser.gz
1  [nsubj(one-3, This-1), cop(one-3, is-2), root(ROOT-0, one-3), prep(one-3, of-4), det(p
2  [root(ROOT-0, -1), det(place-3, The-2), nsubj(clean,-5, place-3), cop(clean,-5, is-4),
3  [root(ROOT-0, -1), nsubj(take-4, I-2), advmod(take-4, usually-3), ccomp(-1, take-4), p
4  [root(ROOT-0, -1), prep(Lechon-9, Among-2), poss(favorites-4, my-3), pobj(Among-2, fav
```

Figure 3 Dependency tree of sentences

The adjectival modifier of a noun phrase (NP) (amod) and the nominal subject (nsubj) dependencies were extracted from the dependency tree. The list of NA pairs in each sentence is appended to a global list of NA pairs for the review. Likewise, parts-of-speech (POS) tagging is performed using Stanford POS Tagger to tag each word according to its part of speech. This will be used to determine the degree of the adjective (positive, comparative, superlative) to give appropriate sentiment weight. Table 1 shows example of POS tags for adjectives in different degrees.

Table 1. Sample POS Tags

| POS Tag | Description | Example |
|---|---|---|
| JJ | adjective | big |
| JJR | adjective, comparative | bigger |
| JJS | adjective, superlative | biggest |

### 3.2 Scoring Algorithm

After getting the list of NA pairs, the system attributes all nouns into five predetermined categories namely: ambiance, cost, food, hygiene, and service by checking its occurrence on a bag-of-words associated with each category.

All noun-adjective pairs which were attributed to the five predetermined categories were processed for polarity. The AFINN lexicon which assigns sentiment weights to NA pairs in the range of -5 to +5. For NA pairs which are not in the AFINN lexicon will be given a default sentiment weight of 1.

Further, a multiplier score is assigned for the degree of adjective, JJ = 1; JJR = 3; JJS = 5. Sample adjectives that match with the tags are enumerated in the form "adjective(POS tag, score)" as follows: bad(JJ:1), best(JJS:5), fresh(JJ:1), great(JJS:5), negative(JJ:1), new(JJ:1), strange(JJ:1).

The final weight of the NA pair is computed as the product of the sentiment weight (sw) based on the AFINN lexicon and the multiplier score representing the degree of the adjective (deg). The rating per category for one review is computed as:

$$rating(X) = \frac{\sum_{i=0}^{n}(sw_i * deg_i)}{n} \qquad (1)$$

where n is the count of distinct reviews from customers and rating(X) is the computed aspect rating per individual review.

$$\overline{rating(X)} = \frac{\sum_{c=0}^{n} rating(X)_c}{n} \qquad (2)$$

International Journal of Computer Science & Information Technology (IJCSIT) Vol 10, No 6, December 2018

The average of the overall aspect rating represents the restaurant star rating. The rating adjusts as new reviews are added. The resulting rating computation would yield a number in the range [-15, +15]. In order to convert computed score to a number in the range [1,5] representing star rating, the affine transformation is applied. Affine transformation maintains the property of order and the relative distances between scale values [8] using the following formula:

$$y = (x - a)\frac{d-c}{b-a} + c \qquad (3)$$

where x is the original computed value in the range [a,b] and y is the transformed value in the range [c,d]. The transformed score is used to assign the overall restaurant star rating. The overall aspect rating is the average of the individual review aspect rating. Figure 4 shows the overall computational process.

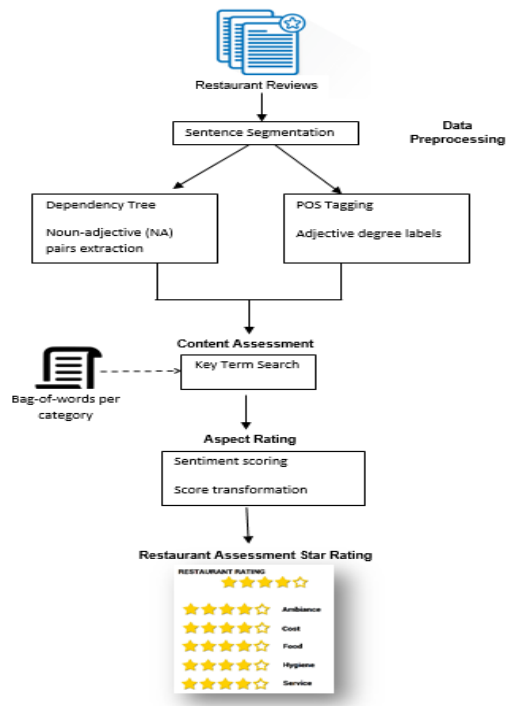

Figure 4. Conceptual Diagram

## 3.3 Word Cloud Generation

A word cloud is a visual representation of frequently used terms in a collection of text. The collective customer reviews made for each restaurant will serve as the text corpus.

Unstructured text underwent data pre-processing such as removal of stop words, punctuation marks and numbers and extra white spaces and converted to lowercase for uniformity in the presentation. A document term matrix was derived the pre-processing stage which is a list of distinct words with its corresponding frequency count for each review as shown in Table 2.

Table 2. Frequently occurring words in the review

| Word | Frequency |
|---|---|
| lechon | 122 |
| good | 38 |
| cebu | 34 |
| place | 25 |
| spicy | 20 |
| best | 19 |
| food | 18 |
| one | 17 |
| taste | 15 |
| like | 14 |

R libraries wordcloud2 and tm were used to generate the word cloud. Rserve facilitates the execution of R scripts and writes the image file of the word cloud into the web server home directory. The word cloud is updated each time a new review is added. Figure 5 shows a sample of a word cloud. The size and the thickness of the word appearing in the image reflect the most frequently occurring descriptor of the restaurant.

Figure 5 Generated wordcloud

## 4. RESULTS AND DISCUSSIONS

The system is weakly supervised using associated words fed to the system. Table 3 shows the list of associated words for each aspect with a total count of 2362.

Table 3. Associated words for each aspect

| Aspect | Associated words | Word count |
|---|---|---|
| ambience | atmosphere, vibe, mood, surroundings… | 168 |
| cost | price, expense, fee, value, pay, budget… | 590 |
| food | meal, gourmet, rice, chicken, pork.. | 640 |
| hygiene | clean, sanitation, tidiness, disposal,.. | 283 |
| service | delivery, care, assistance, employee .. | 681 |

One of the limitations of the application is that it can only assess text reviews written in the English language. Further, the system made use of the lexicon-based sentiment scoring using AFINN. The AFINN lexicon is limited only to 2477 words and therefore may not be able to assign scores to noun-adjective pairs which are not included in the library.

## 4.1 User Interface

Web and mobile client applications were developed to provide an interface for the user to contribute reviews on dining experience (food trip) on a certain restaurant. Customer specifies food trip details including restaurant name, menu ordered, and text review of the experience. Figure 6 shows the user interface used to gather customer feedback through crowdsourcing.

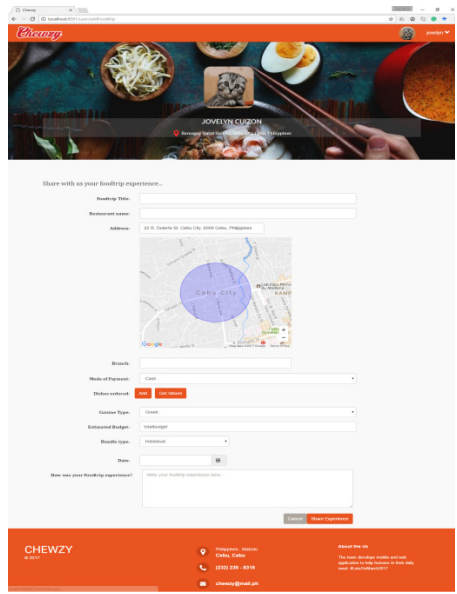

Figure 6. Foodtrip Entry Form

Data contributed by users are processed to automatically assign rating to each pre-identified categories. Each user review contributes to the overall rating of the restaurant and the specific categories evaluated such as ambiance, cost, food, hygiene and service. Figure 7 shows a screenshot of the restaurant profile page which contains system-generated rating.

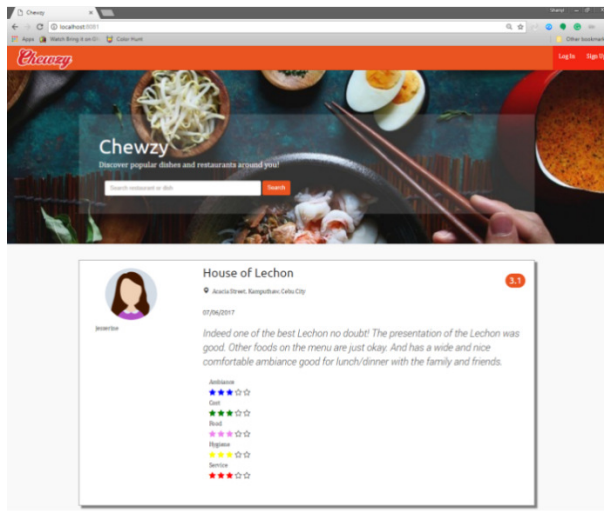

Figure 7 Customer Review with Feature Ratings

## 4.2 Accuracy Testing

The respondents were made to rate each aspect (ambiance, cost, food, hygiene, service) based on a given text reviews which are similar to the ones fed to the system. The experimental value is the aspect rating assigned by the system while theoretical value is the rating assigned through human interpretation. Figure 8 depicts the difference in human and the system generated aspect rating when tested with five (5) distinct text reviews.

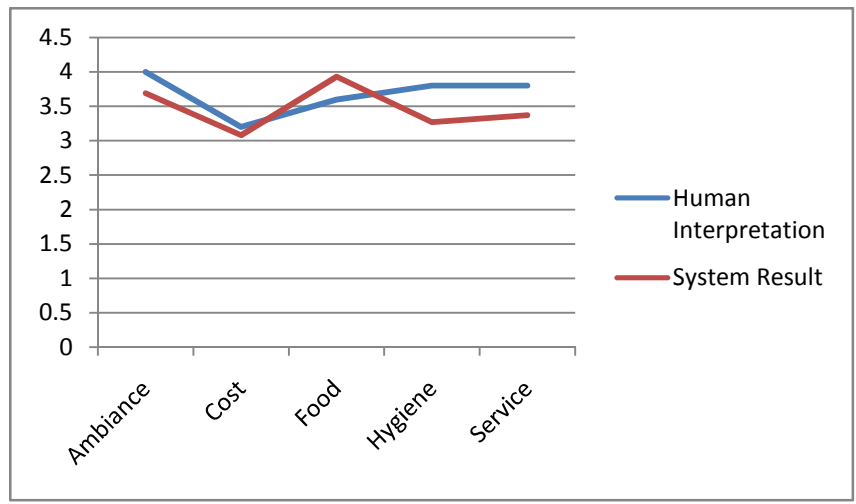

Figure 8. Aspect rating for using human interpretation and system result

In order to quantify the degree of closeness between the system-assigned feature rating and the manually assigned rating based on a given text review, experimental error is derived. The percent error is the ratio of the error to the actual value multiplied by 100. Table 4 shows mean percent error result for each aspect.

$$\text{percent error} = \frac{|\text{experimental value} - \text{theoretical value}|}{\text{theoretical value}} \times 100 \qquad (4)$$

Table 4. Mean percent error per aspect

| Aspect | Human Interpretation | System Result | %error | Point Difference |
|---|---|---|---|---|
| Ambiance | 4.0 | 3.69 | 7.75 | 0.39 |
| Cost | 3.2 | 3.08 | 3.75 | 0.19 |
| Food | 3.6 | 3.93 | 9.16 | 0.46 |
| Hygiene | 3.8 | 3.27 | 13.94 | 0.70 |
| Service | 3.8 | 3.37 | 11.32 | 0.57 |

The system generated some difference in the ratings on each aspect compared to the manually assigned rating. The difference may be influenced by the fact that the human evaluator used whole number values from 1 to 5 to assess while the system generated rating can result to a real number. Cost is the aspect which yields the least percent error because associated words related to cost are unambiguous and plenty which may not be true for hygiene. The point difference in all aspects is lesser than 1 which reflects that the system generated result is generally acceptable.

## 5. CONCLUSIONS

The development of the application is intended to make use of unstructured text to extract relevant descriptors and assessment ratings for user-identified features through natural language processing (NLP). The system could be built on top of existing customer review platform to provide automatic rates of predefined aspects of products and services evaluated.

## REFERENCES

[1] C. C. Aggarwal and C. Zhai, Mining Text Data, Springer Science & Business Media, 2012.

[2] L. Jack and Y. Tsai, "Using Text Mining of Amazon Reviews to Explore," in The 2015 International Conference on Data Mining, Las Vegas, Nevada, USA, 2015.

[3] F. V. Ordenes, ,. B. Theodoulidis, J. Burton, T. Gruber and M. Zaki, "Analyzing Customer Experience Feedback Using Text Mining A Linguistics-Based Approach," Journal of Service Research, vol. 17, no. 4, 2014.

[4] V. Suresh, S. Roohi and M. Eirinaki, "Aspect-Based Opinion Mining and Recommendation System for Restaurant Reviews," in ACM RecSys'14, 2014.

[5] M. Hu and B. Liu, "Mining Opinion Features in Customer Reviews," AAAI, vol. 4, no. 4, 2004.

[6] G. Somprasertsri and P. Lalitrojwong, "Mining Feature-Opinion in Online Customer Reviews for," Journal of Universal Computer Science,, vol. 16, no. 6, pp. 938-955, 2010.

[7] M. Hu and B. Liu, Mining and Summarizing Customer Reviews, in KDD '04 : Proceedings of the tenth ACM SIGKDD international conference on Knowledge discovery and data mining, pp. 168–177, New York, NY, USA, 2004, ACM.

[8] A. Gupta, T. Tenneti and A. Gupta. Sentiment based Summarization of Restaurant Reviews, June 3, 2009.

[9] Chinsa T C, Shibily J. Aspect based Opinion Mining from Restaurant Reviews, International Journal of Computer Applications (0975 – 8887). Advanced Computing and Communication Techniques for High Performance Applications (ICACCTHPA-2014) , 2014.

[10] Titov, Ivan, and Ryan McDonald. "A joint model of text and aspect ratings for sentiment summarization." proceedings of ACL-08: HLT,308-316, 2008

[11] Samaneh Moghaddam and Martin Ester.. Opinion digger: an unsupervised opinion miner from unstructured product reviews. In Proceedings of the 19th ACM international conference on Information and knowledge management (CIKM '10). ACM, New York, NY, USA, 1825-1828. 2010 DOI=http://dx.doi.org/10.1145/1871437.1871739

[12] I. B. Weiner, Handbook of Psychology, Research Methods in Psychology, John Wiley & Sons,, 2003.

## Authors

**Jovelyn Cuizon** is an assistant professor at University of San Jose- Recoletos. She is the academic head for the Computer Science department of the same university. She graduated Master of Science in Information Technology and Doctor in Management in 2004 and 2018 respectively.

**Jesserine Lopez** graduated in 2017 with Bachelor of Science in Computer Science from University of San Jose-Recoletos, Cebu City, Philippines. She is currently a software engineer at Accenture.

**Danica Rose Jones** graduated in 2017 with Bachelor of Science in Computer Science from University of San Jose-Recoletos, Cebu City, Philippines. She is currently a software engineer at Advanced World Systems.